\pdfoutput=1
\documentclass[11pt]{article}

\usepackage[final]{acl}

\usepackage{times}
\usepackage{latexsym}
\usepackage[T1]{fontenc}
\usepackage[utf8]{inputenc}
\usepackage{microtype}
\IfFileExists{inconsolata.sty}{\usepackage{inconsolata}}{}
\usepackage{graphicx}
\usepackage{amsmath}
\usepackage{amssymb}
\usepackage{booktabs}
\usepackage{multirow}
\usepackage{xcolor}
\usepackage{enumitem}
\usepackage{pgfplots}
\pgfplotsset{compat=1.16}
\usepgfplotslibrary{groupplots}

\newcommand{\cmg}{CMG}

\newcommand{\pskip}{p_{\textsc{skip}}}
\newcommand{\puse}{p_{\textsc{use}}}
\newcommand{\yskip}{y_{\textsc{skip}}}
\newcommand{\yuse}{y_{\textsc{use}}}
\newcommand{\SKIP}{\textsc{skip}}
\newcommand{\USE}{\textsc{use}}

\title{SEDM: Selective Self-Evolving Distributed Memory for Agents}

\author{
 \textbf{Yichen Liu}\textsuperscript{\normalfont 1*},
 \textbf{Anna Min}\textsuperscript{\normalfont 2*},
  \textbf{Yinzhi Wang}\textsuperscript{\normalfont 3*},
 \textbf{Yuhang Yao}\textsuperscript{\normalfont 4},
 \textbf{Yichen Han}\textsuperscript{\normalfont 5},
 \textbf{Rui Zhang}\textsuperscript{\normalfont 6}, \\
 \textbf{Jiantao He}\textsuperscript{5},
 \textbf{Zhiheng Li}\textsuperscript{7},
 \textbf{Yingying Qi}\textsuperscript{8},
 \textbf{Haoran Xu}\textsuperscript{9},
 \textbf{Zhimin Ding}\textsuperscript{10},
 \textbf{Yuxin Tang}\textsuperscript{11}, \\
 \textbf{Qi You}\textsuperscript{5},
 \textbf{Lynn Ai}\textsuperscript{12},
 \textbf{Eric Yang}\textsuperscript{12},
 \textbf{Lewei He}\textsuperscript{5},
 \textbf{Tianyu Shi}\textsuperscript{13\dag} \\
 \textsuperscript{1}University of California, San Diego \quad
 \textsuperscript{2}Tsinghua University \quad
 \textsuperscript{3}Baruch College, City University of New York \\
 \textsuperscript{4}Carnegie Mellon University \quad
 \textsuperscript{5}South China Normal University \quad
 \textsuperscript{6}The Ohio State University\\
 \textsuperscript{7}Independent Researcher \quad
 \textsuperscript{8}Nanjing University of Science and Technology \\
 \textsuperscript{9}Polytechnic Institute of Zhejiang University, Zhejiang University \quad
 \textsuperscript{10}Rice University \\
 \textsuperscript{11}Amazon AGI Foundations \quad
 \textsuperscript{12}Gradient \quad
 \textsuperscript{13}McGill University \\
 \textsuperscript{*}Equal contribution. \quad
 \textsuperscript{\dag}Corresponding author: \texttt{tianyu.shi3@mcgill.ca}
}

\begin{document}
\maketitle

\begin{abstract}
Agents in a society can reuse each other's experience, but indiscriminate memory use increases processing cost and can reduce task success.
We present \textbf{SEDM} (Selective Self-Evolving Distributed Memory), a framework that selects both what agents share and when recipients use it.
On the sender side, SEDM admits memories through paired execution tests, ranks them by relevance and measured utility, and abstracts them for cross-domain reuse.
On the recipient side, \textbf{Calibrated Memory Gating (CMG)} predicts success with and without memory in a single forward pass, learning from paired outcomes including ties.
It injects memory only when the calibrated success gain outweighs a context-cost penalty.
SEDM exceeds the strongest evaluated baseline on LoCoMo open-domain questions by 15.1 F1 points and outperforms G-Memory on FEVER and HotpotQA with 16--32\% fewer prompt tokens.
In a controlled HotpotQA study, CMG achieves success comparable to always using memory with 18\% fewer agent input tokens.
Within the full SEDM pipeline on LoCoMo, CMG outperforms both always-use and no-memory policies for two SEDM memories while using 19--31\% fewer input tokens than always-use.
These results show that selective use can recover value from shared memories that hurt performance when injected indiscriminately.
\end{abstract}

\section{Introduction}
\label{sec:introduction}

\begin{figure*}[t]
    \centering
    \includegraphics[width=\textwidth]{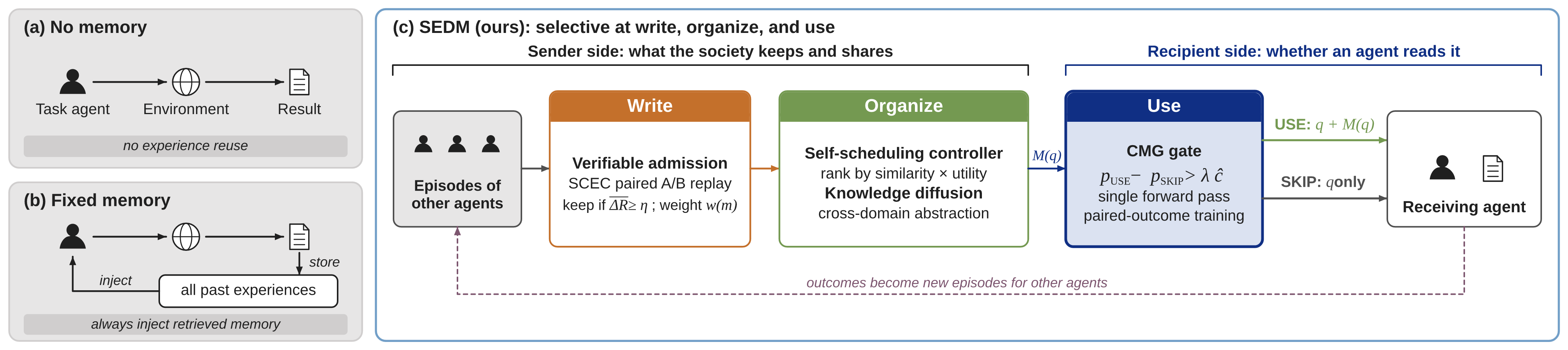}
    \caption{\textbf{Memory strategies for agent societies.} (a) Without memory, agents do not reuse experience. (b) With a fixed memory, all past experiences are stored and retrieved memory is always injected. (c) SEDM is selective on the sender side, through verifiable admission (write) and utility-guided organization, and on the recipient side, where the CMG gate decides whether the receiving agent uses the retrieved memories $M(q)$.}
    \label{fig:overview}
\end{figure*}

Language model agents increasingly operate in societies, dividing work and reusing experience acquired by others \citep{park2023generativeagents, altera2024projectsid, zhang2025gmemorytracinghierarchicalmemory}. Shared memory lets evidence or a solution from one agent inform another agent's decisions. Yet a memory that helped its source task may be redundant, misleading, or too costly for the receiving agent's current task. Society memory must therefore decide both which experiences are worth sharing and when they are worth using (Figure~\ref{fig:overview}).

Existing memory systems organize experience through retrieval, context management, and evolving structures \citep{zhong2024memorybank, packer2024memgptllmsoperatingsystems, xu2025amemagenticmemoryllm, zhang2025gmemorytracinghierarchicalmemory}. But relevance and past utility do not measure the expected improvement over running the current task without memory: a memory can be relevant, leave the agent's chance of success unchanged, and still add context cost. Learning this decision from outcomes is also hard, because winner-only supervision discards ties, where using and skipping memory both succeed or both fail. Ties are 78\% of our HotpotQA training pairs and most of our LoCoMo pairs, yet still carry information about success under both actions.

We present \textbf{SEDM} (Selective Self-Evolving Distributed Memory), which uses execution outcomes to judge memory twice: whether an experience merits retention, and whether a retrieved experience is worth using now. On the sender side, SEDM tests candidate memories with paired replays of Self-Contained Execution Contexts (SCECs), schedules stored items by relevance and measured utility, and abstracts them for cross-domain reuse. On the recipient side, \textbf{Calibrated Memory Gating (CMG)} predicts success with and without retrieved memory in a single forward pass, learns from paired outcomes including ties, and injects memory when the calibrated gain exceeds a context-cost penalty, so the trade-off can change without retraining.

Under the original admission configuration, SEDM exceeds the strongest evaluated baseline on LoCoMo open-domain questions by 15.1 F1 points and outperforms G-Memory on FEVER and HotpotQA with 16--32\% fewer prompt tokens. We evaluate CMG where memory helps on average and where indiscriminate use hurts. In a controlled HotpotQA study, CMG approaches always-use success (32.9\% vs.\ 33.3\%) with 18\% fewer agent input tokens. Within the full SEDM pipeline on LoCoMo, with held-out admission, CMG achieves higher observed success than both fixed policies across two memory configurations, with 19--31\% fewer input tokens than always-use, recovering value even when indiscriminate injection hurts.

Our contributions are:
\begin{itemize}[topsep=2pt, itemsep=1pt, parsep=0pt]
    \item We propose SEDM, a shared-memory framework for agent societies that couples sender-side admission and organization with recipient-side decisions on memory use.
    \item We develop CMG, a single-pass gate that learns action-specific success probabilities from paired outcomes, including ties, and weighs calibrated gains against context cost.
    \item We evaluate on LoCoMo, FEVER, and HotpotQA, showing that CMG lowers the cost of helpful memory and recovers value from memory that hurts when always injected.
\end{itemize}

\section{Related Work}
\label{sec:related}

\paragraph{Agent societies and shared memory.}
Generative agents simulate small communities that remember and reflect \citep{park2023generativeagents}, and many-agent simulations scale this to societies with roles, rules, and culture \citep{altera2024projectsid}. Multi-agent systems increasingly share memory across agents \citep{zhang2025gmemorytracinghierarchicalmemory}. Most of this work focuses on what the society stores and passes on. SEDM addresses both sides: what the society keeps, through verified admission and utility-guided management, and whether a receiving agent should use what it is given.

\begin{figure*}[t]
    \centering
    \makebox[\textwidth][c]{\includegraphics[width=1.02\textwidth]{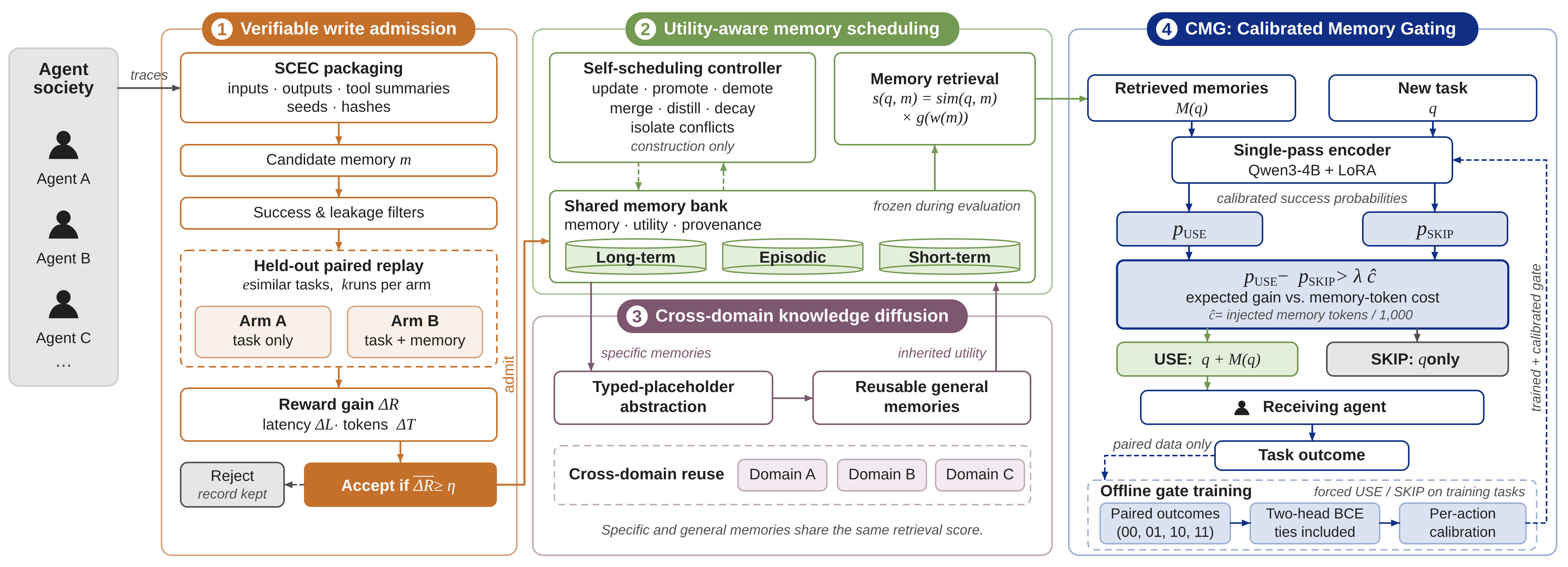}}
    \caption{\textbf{Full SEDM architecture.} SEDM integrates SCEC-based verifiable write admission, utility-aware memory scheduling, cross-domain knowledge diffusion, and calibrated memory gating (CMG), which decides whether to USE or SKIP retrieved memories.}
    \label{fig:architecture}
\end{figure*}
\vspace{0.8cm}
\paragraph{Memory for language agents.}
Agents store and reuse experience in many forms: verbal reflections \citep{shinn2023reflexion}, skill libraries \citep{wang2023voyager}, episodic streams \citep{park2023generativeagents}, OS-style context management \citep{packer2024memgptllmsoperatingsystems}, decaying stores \citep{zhong2024memorybank}, linked notes \citep{xu2025amemagenticmemoryllm}, hierarchical multi-agent memory \citep{zhang2025gmemorytracinghierarchicalmemory}, and production memory layers \citep{chhikara2025mem0buildingproductionreadyai}; see \citet{zhang2024surveymemory} for a survey. These systems mainly decide \emph{what} to store and retrieve. \cmg{} addresses the last step: whether the retrieved memories should be shown to the agent at all.

\paragraph{Adaptive retrieval.}
Several works decide when a language model should retrieve: from a learned reflection token \citep{asai2024selfrag}, from generation confidence \citep{jiang2023flare}, from entity popularity \citep{mallen2023popqa}, or from a question-complexity classifier \citep{jeong2024adaptiverag}. These methods target document retrieval for a single generation. We target agent memory injected into a multi-step agent, supervise the gate with the agent's own outcomes under both actions, and make the token cost an explicit part of the decision.

\paragraph{Calibration and selective decisions.}
Temperature scaling is a simple and strong way to calibrate neural classifiers \citep{guo2017calibration}, and calibration is commonly measured with NLL, Brier score, and expected calibration error \citep{naeini2015ece}. Language models can estimate their own correctness \citep{kadavath2022language}, and selective prediction trades coverage against risk \citep{geifman2017selective}. \cmg{} applies these ideas to a two-action decision: it estimates the success probability of each action and trades the predicted gain against a cost.

\section{Methodology}
\label{sec:method}

\subsection{System Overview}
\label{sec:overview}

SEDM is a shared memory for a society of agents. Agents solve tasks, and what they learn becomes candidate memory for other agents. A shared memory of this kind can easily fill up with items that are noisy, redundant, or useful only for the task they came from. SEDM therefore makes memory \emph{selective} at three points. (1) \textbf{Write.} A candidate is admitted only if a replayed A/B test shows that it does not hurt (\S\ref{sec:admission}). (2) \textbf{Organize.} A controller ranks, consolidates, and retires admitted items by their measured utility, and abstracts them for reuse across domains (\S\ref{sec:controller}, \S\ref{sec:diffusion}). (3) \textbf{Use.} When an agent receives retrieved memories for a new task, a calibrated gate decides whether to inject them at all (\S\ref{sec:cmg}). The first two steps work on the sender side: they decide what the society keeps and shares. The third step works on the recipient side: it decides whether a receiving agent should read what it is given. Figure~\ref{fig:architecture} shows the full pipeline and the components of each step.

Throughout the paper, $q$ is a task (for example, a question), $m$ is a memory item, and $\mathcal{F}$ is the agent. For an agent output $o$, $R(o)$ is its task reward, $L(o)$ its latency, and $T(o)$ its token usage.

\subsection{SCEC-based Verifiable Write Admission}
\label{sec:admission}

\paragraph{Self-contained packaging and replay.}
Each task run is packaged as a \emph{Self-Contained Execution Context} (SCEC). An SCEC stores the task input, the agent output, summaries of tool calls, random seeds, and hashes of the configuration. Because external tool calls are stored as summaries, an SCEC can be replayed without the original environment. SCECs can run as independent jobs on distributed workers, and only aggregated statistics, integrity hashes, and version stamps are sent back. This makes admission reproducible and easy to run in parallel.

\paragraph{Paired A/B test.}
From each SCEC we extract one candidate memory $m$ and test it on a set of tasks $Q(m)$. In the original SEDM configuration, $Q(m)$ is the source task itself, whose SCEC is replayed with and without $m$. Because the agent has already seen the source task, this can overestimate the utility of $m$. Our stricter \emph{held-out} configuration therefore uses the $e$ held-out admission tasks that are most similar to $m$, and replays their SCECs instead. For each $q \in Q(m)$, the control run sees only the task, and the treatment run also sees the memory:
\begin{equation}
o_A = \mathcal{F}\big(f(q)\big), \qquad o_B = \mathcal{F}\big(f(q; m)\big),
\end{equation}
where $f$ builds the prompt. We record the changes in reward, latency, and tokens, $\Delta R = R(o_B) - R(o_A)$, $\Delta L$, and $\Delta T$. Each arm is run $k$ times, and the deltas are averaged over the runs and over the tasks in $Q(m)$, which reduces the noise of a single comparison. The admission score trades the reward change against its cost:
\begin{align}
S(m) &= \overline{\Delta R}(m) - \lambda_L \overline{\Delta L}(m) - \lambda_T \overline{\Delta T}(m), \\
\mathrm{accept}(m) &\iff S(m) \ge \eta ,
\end{align}
with $\lambda_L, \lambda_T \ge 0$. In the held-out configuration we set $\lambda_L = \lambda_T = 0$, so admission depends on the reward change alone. The latency and token changes are stored with the item, together with its provenance (hashes, seeds, versions). In the held-out configuration, two filters run before the test: a candidate is dropped if its source task was not solved, and if it contains the reference answer of a task it is tested on, since it would then pass only by leaking the answer.

\paragraph{Weight initialization.}
In the original configuration, an admitted item starts with weight $w_0(m) = \max\{0, S(m)\}$. In the held-out configuration, every tested candidate receives a weight record. It starts from a small base weight $w_0$ and moves up or down by a bounded step proportional to $\overline{\Delta R}$, so a memory that helped starts above the base weight, a memory that hurt starts below it, and a tie keeps the base weight (exact rule in Appendix~\ref{app:sedm}). Because $w_0 > 0$, tied items can still be retrieved and collect further evidence later. Rejected candidates are never retrieved; their records are kept only for auditing and conflict isolation.

\subsection{Self-Scheduling Memory Controller}
\label{sec:controller}

Admission tests a memory once, on a few similar tasks. The controller keeps the memory organized as more evidence arrives, and it does so without calling an LLM for every query.

\paragraph{Retrieval-time scheduling.}
For a new task $q$, each stored item is scored by its cosine similarity to the task, scaled by a smooth, increasing function of its weight:
\begin{equation}
s(q, m) = \mathrm{sim}(q, m) \cdot g\big(w(m)\big),
\end{equation}
where the original configuration uses $g(w) = w$, and the held-out configuration uses the bounded form $g(w) = 1.1 + 0.4\tanh(\tfrac{1}{2}\ln w)$, which lies in $(0.7, 1.23]$ for $w \in (0, 2]$. The weight can therefore reorder items of similar relevance without overriding relevance itself. The candidates are then re-ranked by tier (long-term before episodic before short-term) and, within a tier, by weight, and the top items form the retrieved set $M(q)$.

\paragraph{Consolidation and progressive evolution.}
The controller runs periodically and once more at the end of a run. In each cycle it (1) applies new A/B evidence to item weights and counts each use (Appendix~\ref{app:sedm} gives the update rule of each configuration); (2) promotes items with high weight and repeated use to the episodic and then the long-term tier, and demotes items whose weight falls; (3) merges near-duplicate items from different SCECs, distills repeated patterns into new \emph{distilled} items, and isolates items whose tests lower the reward below a conflict threshold, so that they are never retrieved again; and (4) decays the weights of items that are not used and removes items whose weight falls below a minimum. Frequently helpful items thus rise to the long-term tier, while unused or harmful items fade out of retrieval. Thresholds and decay constants are listed in Appendix~\ref{app:sedm}.

\subsection{Cross-Domain Knowledge Diffusion}
\label{sec:diffusion}

Memories are also made portable across task domains. For each admitted item, a rule-based abstraction replaces specific entities with typed placeholders and produces a general form, $m_{\text{general}} = \mathrm{Abstract}(m_{\text{specific}})$. The general form inherits a conservative weight, a fixed fraction of the weight of the specific item. Both forms compete under the same retrieval score $s(q,m)$: in-domain queries tend to retrieve the specific form, and queries from other domains tend to retrieve the general form.

\subsection{Selective Memory Use: Calibrated Memory Gating}
\label{sec:cmg}

Admission and the controller decide what the society keeps. They do not decide whether a particular agent should read the retrieved memories for its current task. A memory that passed admission on similar tasks can still be unhelpful, misleading, or simply too costly for a new task. Unlike a generative chain-of-thought (CoT) gate, which is slow and gives no confidence to weigh against cost, we propose \emph{Calibrated Memory Gating} (\cmg{}), a single-pass gate that predicts the outcome of both actions and trades the predicted gain against cost.

\paragraph{Gating decision.}
For a task $q$, the controller returns the set $M(q)$. The gate chooses an action $a \in \{\SKIP, \USE\}$. Under \SKIP{} the agent sees only the task; under \USE{} it also sees all of $M(q)$. Let $y_a(q) \in \{0,1\}$ be whether the agent succeeds under action $a$, and $c_a(q)$ the cost of that action in tokens. We want a policy that maximizes success minus $\lambda$ times cost, where $\lambda \ge 0$ is the price of one thousand tokens in units of success probability.

\paragraph{Paired execution data.}
Like admission, \cmg{} learns from paired executions, but on the \emph{current} task with the whole \emph{retrieved set}. For each training task we run the agent twice, without and with $M(q)$, and record $(\yskip, \yuse)$. Groups 01 (only \USE{} succeeds) and 10 (only \SKIP{} succeeds) say which action is better; the ties 00 and 11 still show how likely the agent is to succeed. Because evaluation tasks are also run under both actions, any gating policy can be scored exactly on the same executions.

\paragraph{Model and paired outcome loss.}
The gate input contains the task and the retrieved memories, ending with the cue ``\texttt{Decision (skip or use):}''; it never contains answers, scores, or execution traces. A Qwen3-4B model \citep{yang2025qwen3} with LoRA adapters \citep{hu2022lora} encodes this text once. A linear head on the last hidden state $h$ gives two logits $z = Wh + b$ and two independent probabilities, $\pskip = \sigma(z_1)$ and $\puse = \sigma(z_2)$; they need not sum to one, because both actions can succeed or both can fail. We initialize $W$ to zero and $b$ to the log-odds of the training success rate of each action. Each probability is trained on the outcome of its own action:
\begin{equation}
\mathcal{L} = \mathrm{BCE}(\pskip, \yskip) + \mathrm{BCE}(\puse, \yuse).
\end{equation}
All four groups are used, including ties. Ties teach the gate how hard a task is; groups 01 and 10 teach it when memory changes the outcome. A winner label on non-tie tasks only would discard 78\% of the training tasks on HotpotQA and most of them on LoCoMo.

\paragraph{Calibration.}
Raw probabilities from a fine-tuned model are often overconfident. After training, we fit one temperature per action on a separate calibration split by minimizing NLL \citep{guo2017calibration}: $p_a = \sigma(z_a / T_a)$. Temperature does not change the ranking of a single head, but it changes the size of $\puse - \pskip$, and therefore the decision.

\paragraph{Cost-aware decision.}
\cmg{} uses memory if and only if
\begin{equation}
\puse(q) - \pskip(q) > \lambda\,\big(\hat c_{\USE}(q) - \hat c_{\SKIP}(q)\big),
\label{eq:rule}
\end{equation}
where $\hat c$ is a cost estimate known before execution: the length of the injected memory text in thousands of tokens, and $\hat c_{\SKIP}=0$. With $\lambda=0$ the gate uses memory whenever it predicts a gain; larger $\lambda$ asks for a larger gain per token. We pick $\lambda$ from $\{0, 0.05, 0.1, 0.2, 0.5, 1\}$ on the validation split and also report the full curve. The rule mirrors admission: admission asks whether a memory improved similar tasks enough to keep it, and the gate asks whether the retrieved memories are expected to improve the current task enough to justify their cost.

\section{Experiments}
\label{sec:experiments}

\subsection{Experimental Setup}
\label{sec:setup}

\paragraph{Datasets and metrics.}
We evaluate on LoCoMo \citep{maharana2024evaluating} (long multi-session conversations with five question types; token F1 and BLEU-1), FEVER \citep{thorne2018fever} (fact verification; accuracy), and HotpotQA \citep{yang2018hotpotqa} (multi-hop question answering; exact match). For FEVER and HotpotQA we also report prompt and completion tokens as a measure of cost.

\paragraph{Models and baselines.}
All agents use gpt-4o-mini, and dense retrieval uses all-MiniLM-L6-v2 \citep{reimers2019sentencebert,wang2020minilm}. On LoCoMo we compare with nine memory systems (Table~\ref{tab:locomo-main}); their results are those of the original SEDM study, several of which are taken from prior work \citep{xu2025amemagenticmemoryllm}, so their memory budgets may differ. On FEVER and HotpotQA we compare with no memory and with G-Memory under the same memory budget. Tables~\ref{tab:locomo-main}--\ref{tab:cross} use the original SEDM configuration (\S\ref{sec:admission}), while \S\ref{sec:locomo} uses the stricter held-out configuration.

\paragraph{Selective-use protocol.}
To evaluate \cmg{}, every task is executed both without and with the retrieved memories, so that any gating policy can be scored exactly on the same runs (\S\ref{sec:cmg}). We use a controlled HotpotQA study with a fixed ReAct agent \citep{yao2023react}, where a task succeeds if the answer reaches token F1 $\ge 0.5$ (Appendix~\ref{app:hotpot}), and the full SEDM pipeline on LoCoMo (\S\ref{sec:locomo}). Both use their own splits and success criteria, so their numbers are not comparable to Tables~\ref{tab:locomo-main}--\ref{tab:cross}. All gates use Qwen3-4B with LoRA (rank 16) on one NVIDIA A40; \cmg{} is trained with three seeds.

\subsection{Main Results on LoCoMo}
\label{sec:main}

\begin{table*}[t]
\centering
\small
\setlength{\tabcolsep}{5pt}
\begin{tabular}{@{}l|cc|cc|cc|cc|cc@{}}
\toprule
\multirow{2}{*}{Method} & \multicolumn{2}{c|}{Single Hop} & \multicolumn{2}{c|}{Multi-Hop} & \multicolumn{2}{c|}{Open Domain} & \multicolumn{2}{c|}{Temporal} & \multicolumn{2}{c}{Adversarial} \\
 & F1 $\uparrow$ & B1 $\uparrow$ & F1 $\uparrow$ & B1 $\uparrow$ & F1 $\uparrow$ & B1 $\uparrow$ & F1 $\uparrow$ & B1 $\uparrow$ & F1 $\uparrow$ & B1 $\uparrow$ \\
\midrule
LoCoMo \citep{maharana2024evaluating} & 40.4 & 29.1 & 25.0 & 19.8 & 12.0 & 11.2 & 18.4 & 14.8 & \textbf{69.2} & \textbf{68.8} \\
ReadAgent \citep{lee2024readagent} & 9.7 & 7.7 & 9.2 & 6.5 & 5.3 & 5.1 & 12.6 & 8.9 & 9.8 & 9.0 \\
MemoryBank \citep{zhong2024memorybank} & 6.6 & 5.2 & 5.0 & 4.8 & 5.6 & 5.9 & 9.7 & 7.0 & 7.4 & 6.5 \\
MemGPT \citep{packer2024memgptllmsoperatingsystems} & 41.0 & 34.3 & 26.7 & 17.7 & 9.2 & 7.4 & 25.5 & 19.4 & 43.2 & 42.7 \\
A-Mem \citep{xu2025amemagenticmemoryllm} & 44.7 & 37.1 & 27.0 & 20.1 & 12.1 & 12.0 & \textbf{45.9} & \textbf{36.7} & 50.0 & 49.5 \\
Zep \citep{rasmussen2025zep} & 26.7 & 18.4 & 30.2 & 17.2 & 3.5 & 2.7 & 15.0 & 11.6 & 22.6 & 15.1 \\
LangMem \citep{langmem2025} & 31.6 & 23.9 & 22.4 & 15.2 & 27.8 & 21.5 & 18.7 & 16.0 & 28.3 & 21.3 \\
Mem0 \citep{chhikara2025mem0buildingproductionreadyai} & 34.0 & 24.8 & 27.3 & 18.6 & 26.9 & 21.1 & 18.6 & 13.9 & 30.4 & 22.2 \\
G-Memory \citep{zhang2025gmemorytracinghierarchicalmemory} & \textbf{53.5} & \textbf{44.0} & \textbf{34.6} & \textbf{26.6} & 32.4 & 25.6 & 9.1 & 7.2 & 11.3 & 9.3 \\
\midrule
\textbf{SEDM w/o \cmg{} (ours)} & 51.7 & 37.0 & 33.5 & 24.4 & \textbf{47.5} & \textbf{33.1} & 12.1 & 9.2 & 12.1 & 9.3 \\
\bottomrule
\end{tabular}
\caption{Results on LoCoMo. F1: token F1; B1: BLEU-1. Best result per column in bold. SEDM here injects every retrieved memory (no \cmg{}); \cmg{} is evaluated in Sections~\ref{sec:cmg-exp}--\ref{sec:locomo}.}
\label{tab:locomo-main}
\end{table*}
SEDM gives the best open-domain F1 (47.5), 15.1 points above the strongest baseline (G-Memory, 32.4), and is second on multi-hop and single-hop questions (33.5 vs.\ 34.6 and 51.7 vs.\ 53.5; Table~\ref{tab:locomo-main}). SEDM builds on G-Memory retrieval, so G-Memory is close to SEDM without admission and scheduling; SEDM improves on it for three of the five types (temporal $+$3.0, open-domain $+$15.1, adversarial $+$0.8). We attribute the open-domain gain to verified, utility-ranked memories that add useful background knowledge. Temporal and adversarial questions favor other designs, such as linked notes (A-Mem) and refusing unanswerable questions (the LoCoMo baseline).

\subsection{Accuracy, Efficiency, and Ablation on FEVER and HotpotQA}
\label{sec:efficiency}

\begin{table}[t]
\centering
\small
\setlength{\tabcolsep}{3.5pt}
\resizebox{\columnwidth}{!}{%
\begin{tabular}{@{}lccc|ccc@{}}
\toprule
 & \multicolumn{3}{c|}{FEVER} & \multicolumn{3}{c}{HotpotQA} \\
Method & Acc. & Prompt & Compl. & EM & Prompt & Compl. \\
\midrule
No memory & 57 & 1.65M & 24K & 34 & 2.46M & 29K \\
G-Memory & 62 & 3.62M & 109K & 38 & 4.63M & 114K \\
\midrule
\multicolumn{7}{@{}l}{\textit{SEDM w/o \cmg{}, adding components}} \\
\quad + SCEC admission & 64 & 2.19M & 53K & 37 & 3.52M & 52K \\
\quad + self-scheduling & \textbf{66} & 2.47M & 53K & \textbf{39} & 3.88M & 55K \\
\bottomrule
\end{tabular}}
\caption{Accuracy and token cost on FEVER and HotpotQA. Prompt/Compl.: total prompt and completion tokens. SEDM rows add components cumulatively. SEDM here injects every retrieved memory (no \cmg{}); \cmg{} is evaluated in Sections~\ref{sec:cmg-exp}--\ref{sec:locomo}.}
\label{tab:efficiency}
\end{table}

SEDM gives the best score on both tasks (Table~\ref{tab:efficiency}): 66 on FEVER (vs.\ 62 for G-Memory and 57 without memory) and 39 on HotpotQA (vs.\ 38 and 34). It also uses 32\% and 16\% fewer prompt tokens than G-Memory and about half of its completion tokens. Admission alone already uses 40\% fewer prompt tokens than G-Memory on FEVER (2.19M vs.\ 3.62M) and is more accurate, which shows that paired tests keep noisy items out. On HotpotQA, admission alone (37) is still below G-Memory (38); self-scheduling then adds 2 points on both tasks with only a small increase in prompt tokens, because utility-weighted retrieval puts helpful memories first instead of injecting more of them.

\subsection{Cross-Domain Evaluation}
\label{sec:cross}

\begin{table}[t]
\centering
\small
\begin{tabular}{@{}l|ccc@{}}
\toprule
Collect $\downarrow$ / Test $\rightarrow$ & FEVER & HotpotQA & LoCoMo \\
\midrule
No memory & 57 & 34 & -- \\
\midrule
FEVER & \textbf{66} & \textbf{41} & 38.1 \\
HotpotQA & 64 & 39 & \textbf{38.6} \\
LoCoMo & 65 & 34 & 37.6 \\
\bottomrule
\end{tabular}
\caption{Cross-domain evaluation of SEDM w/o \cmg{}. Rows give the dataset on which memory is collected, and columns the dataset on which it is tested (FEVER: accuracy; HotpotQA: EM; LoCoMo: overall token F1). The LoCoMo scores come from the original cross-domain runs and are not directly comparable to the per-type scores in Table~\ref{tab:locomo-main}.}
\label{tab:cross}
\end{table}

Table~\ref{tab:cross} collects memory on one dataset and tests it on another. Memory from FEVER even beats in-domain memory on HotpotQA (41 vs.\ 39), memory from HotpotQA and LoCoMo stays close to in-domain memory on FEVER (64 and 65 vs.\ 66) and well above no memory (57), and on LoCoMo all three sources fall within one point. Only LoCoMo memory does not help HotpotQA (34, equal to no memory), which suggests that dialogue knowledge helps little for multi-hop reasoning over documents. All runs use cross-domain diffusion, so these results show that SEDM memories transfer, but they do not isolate how much of the transfer is due to abstraction.
\vspace{0.5cm}
\subsection{Selective Memory Use with CMG}
\label{sec:cmg-exp}
\vspace{0.3cm}
\paragraph{Controlled study on HotpotQA.}
\label{sec:exp}
The experiments above inject every retrieved memory. We now test whether \cmg{} can decide when to use them, first in a controlled study (Appendix~\ref{app:hotpot}): each memory records one past episode from a separate split (question, verified answer, and supporting sentences), and the five most similar memories are retrieved for each question. Table~\ref{tab:main} compares \cmg{} with fixed policies, a similarity threshold, two gates trained on winner labels of non-tie tasks (a winner head and a single-token SFT gate), and a generative CoT gate on the same backbone. Always using memory raises success from 0.249 to 0.333 but more than doubles the agent's input tokens. Memory turns 73 failures into successes but also 32 successes into failures, so a perfect gate would reach 0.399.

\begin{table*}[t]
\centering
\small
\begin{tabular}{@{}lcccc@{}}
\toprule
Gate & Success [95\% CI] & Injection rate & Agent input tokens & Gate latency (s) \\
\midrule
Always skip (no memory) & 0.249 [0.213, 0.286] & 0\% & 1,629 & -- \\
Always use & 0.333 [0.292, 0.376] & 100\% & 3,736 & -- \\
Similarity threshold ($\tau$ = 0.2) & 0.333 [0.292, 0.376] & 100\% & 3,736 & -- \\
Single-token SFT (winner labels) & 0.333 [0.292, 0.376] & 100\% & 3,736 & 0.24 \\
Winner head (winner labels) & 0.333 [0.292, 0.376] & 100\% & 3,736 & 0.12 \\
CoT gate (generative) & 0.262 [0.225, 0.301] & 9\% & 1,734 & 3.09 \\
\midrule
\cmg{} without calibration, $\lambda=0$ & 0.327 [0.286, 0.370] & 71\% & 3,070 & 0.24 \\
\textbf{\cmg{}, $\lambda=0$ (selected on validation)} & \textbf{0.329 [0.288, 0.372]} & \textbf{69\%} & \textbf{3,047 ($-$18\%)} & \textbf{0.24} \\
\cmg{}, $\lambda=0.05$ & 0.315 [0.276, 0.354] & 40\% & 2,487 ($-$33\%) & 0.24 \\
\cmg{}, $\lambda=0.2$ & 0.282 [0.243, 0.321] & 13\% & 1,856 ($-$50\%) & 0.24 \\
\midrule
Oracle (upper bound) & 0.399 & -- & -- & -- \\
\bottomrule
\end{tabular}
\caption{Gate comparison in the controlled HotpotQA study (test split, 489 questions, seed 42). Token savings are relative to always use. All learned gates share the Qwen3-4B backbone; latency was measured on a shared A40, and the two single-pass heads differ only by server load.}
\label{tab:main}
\end{table*}
\paragraph{\cmg{} keeps the quality and cuts the cost.}
With the validation-selected $\lambda=0$, \cmg{} reaches 0.329, within the confidence interval of always use, while injecting memory on 69\% of the tasks and using 18\% fewer agent input tokens. At $\lambda=0.05$ it injects on 40\%, saves 33\% of the tokens, and keeps 79\% of the gain of always using memory (Figure~\ref{fig:curve}a).

\begin{figure*}[t]
\centering
\begin{tikzpicture}
\begin{groupplot}[
  group style={group size=3 by 1, horizontal sep=1.0cm},
  width=0.355\textwidth, height=0.27\textwidth,
  xlabel={Agent input tokens per task},
  scaled x ticks=false,
  xticklabel style={/pgf/number format/1000 sep={\mathord{,}}},
  yticklabel style={/pgf/number format/fixed,/pgf/number format/precision=2},
  tick label style={font=\scriptsize}, label style={font=\small}, title style={font=\small},
  grid=major, grid style={gray!20},
  legend style={font=\scriptsize, draw=none, fill=white, fill opacity=0.85, text opacity=1},
]
\nextgroupplot[title={(a) HotpotQA, evidence memory}, ylabel={Success rate},
  xmin=1500, xmax=3900, ymin=0.235, ymax=0.41, xtick={2000,3000},
  legend columns=5, legend to name=cmglegend,
  legend style={/tikz/every even column/.append style={column sep=3pt}}]
\addplot[densely dashed, gray, thick] coordinates {(1629,0.249) (3736,0.333)};
\addlegendentry{Random gating}
\addplot[blue!70!black, thick, mark=*, mark size=1.5pt] coordinates {
  (1629,0.249) (1672,0.262) (1856,0.282) (2159,0.288) (2487,0.315) (3047,0.329)};
\addlegendentry{\cmg{}}
\addplot[only marks, mark=o, mark size=2.3pt, thick, red!70!black] coordinates {(1734,0.262)};
\addlegendentry{CoT gate}
\addplot[only marks, mark=square*, mark size=1.7pt, black] coordinates {(3736,0.333)};
\addlegendentry{Always use memory}
\addplot[only marks, mark=triangle*, mark size=2.2pt, black!60] coordinates {(1629,0.249)};
\addlegendentry{No memory}
\addplot[dotted, black] coordinates {(1500,0.399) (3900,0.399)};
\node[font=\tiny, anchor=south west] at (axis cs:1520,0.399) {Oracle 0.399};
\node[font=\tiny, anchor=south east] at (axis cs:3047,0.331) {$\lambda{=}0$};
\node[font=\tiny, anchor=south east] at (axis cs:2487,0.317) {$0.05$};
\node[font=\tiny, anchor=south east] at (axis cs:1856,0.284) {$0.2$};
\nextgroupplot[title={(b) LoCoMo, admission only},
  xmin=1400, xmax=2900, ymin=0.505, ymax=0.62, xtick={1500,2000,2500}]
\addplot[densely dashed, gray, thick] coordinates {(1476,0.544) (2790,0.539)};
\addplot[blue!70!black, thick, mark=*, mark size=1.5pt] coordinates {
  (1476,0.544) (1503,0.556) (1825,0.568) (2263,0.556)};
\addplot[only marks, mark=square*, mark size=1.7pt, black] coordinates {(2790,0.539)};
\addplot[only marks, mark=triangle*, mark size=2.2pt, black!60] coordinates {(1476,0.544)};
\addplot[dotted, black] coordinates {(1400,0.606) (2900,0.606)};
\node[font=\tiny, anchor=south west] at (axis cs:1420,0.606) {Oracle 0.606};
\node[font=\tiny, anchor=south] at (axis cs:2263,0.558) {$\lambda{=}0$};
\node[font=\tiny, anchor=south] at (axis cs:1825,0.570) {$0.05$};
\node[font=\tiny, anchor=north west] at (axis cs:1476,0.541) {No memory};
\nextgroupplot[title={(c) LoCoMo, admission + controller},
  xmin=1400, xmax=2900, ymin=0.505, ymax=0.62, xtick={1500,2000,2500}]
\addplot[densely dashed, gray, thick] coordinates {(1476,0.534) (2781,0.518)};
\addplot[blue!70!black, thick, mark=*, mark size=1.5pt] coordinates {
  (1476,0.534) (1558,0.537) (1915,0.546)};
\addplot[only marks, mark=square*, mark size=1.7pt, black] coordinates {(2781,0.518)};
\addplot[only marks, mark=triangle*, mark size=2.2pt, black!60] coordinates {(1476,0.534)};
\addplot[dotted, black] coordinates {(1400,0.596) (2900,0.596)};
\node[font=\tiny, anchor=south west] at (axis cs:1420,0.596) {Oracle 0.596};
\node[font=\tiny, anchor=south] at (axis cs:1915,0.548) {$\lambda{=}0$};
\node[font=\tiny, anchor=north west] at (axis cs:1476,0.531) {No memory};
\end{groupplot}
\end{tikzpicture}\\[2pt]
\pgfplotslegendfromname{cmglegend}
\caption{Quality--cost trade-off of \cmg{}. Each blue curve sweeps the price $\lambda$ from large (left, memory almost never used) to $\lambda=0$ (right). The dashed line is the expected result of random gating between no memory and always using memory; black squares mark always using memory. (a) HotpotQA test split, seed 42, with the generative CoT gate for reference. (b, c) LoCoMo test conversation, mean of three seeds, for the two SEDM memories. In all three settings the \cmg{} curve lies above random gating.}
\label{fig:curve}
\end{figure*}
\vspace{0.5cm}
\paragraph{\cmg{} selects well, not just less.} Random gating at injection rate $u$ scores $0.249 + 0.084u$ in expectation. \cmg{} lies 1.9--3.2 points above this line at every point of the curve, so its savings do not come from simply using memory less often.

\paragraph{The generative CoT gate is slow and too cautious.} It needs 3.09\,s per decision, 13 to 26 times longer than the single-pass gates, injects memory on only 9\% of the tasks, and reaches 0.262, only 0.5 points above random gating at that rate.

\paragraph{Calibration stabilizes decisions.} Across seeds, the raw gate at $\lambda=0$ ranges from 0.305 to 0.339, while the calibrated gate stays between 0.323 and 0.329 (Appendix~\ref{app:hotpot}, Table~\ref{tab:calib}).
\vspace{0.6cm}
\subsection{Selective Use inside the Full SEDM Pipeline}
\label{sec:locomo}

Here the memories that \cmg{} judges are produced, admitted, and managed by SEDM itself. We split LoCoMo by conversation: SEDM builds candidate memories from conversation 0 and admits them with held-out paired tests on conversation 1 (\S\ref{sec:admission}). From the same steps we build two frozen society memories, one with admission only and one in which SEDM also runs its controller (\S\ref{sec:controller}). For each memory, \cmg{} is trained on conversations 3--7 (1,089 questions) and tested on conversation 2 (193 questions); every question is executed through the full pipeline with memory forced off and on (Appendix~\ref{app:locomo}).

\begin{table}[t]
\centering
\small
\setlength{\tabcolsep}{4pt}
\begin{tabular}{@{}lcccc@{}}
\toprule
Setting & Succ. & F1 & Inj. & Tok. \\
\midrule
\multicolumn{5}{@{}l}{\textit{SEDM memory: admission only}} \\
No memory & 0.544 & 0.556 & 0\% & 1,476 \\
SEDM, no gate & 0.539 & 0.564 & 100\% & 2,790 \\
SEDM + \cmg{} (val.\ $\lambda{=}0$) & \textbf{0.556} & \textbf{0.577} & 65\% & 2,263 \\
SEDM + \cmg{} ($\lambda{=}0.05$) & \textbf{0.568} & \textbf{0.579} & 30\% & 1,825 \\
Oracle & 0.606 & -- & -- & -- \\
\midrule
\multicolumn{5}{@{}l}{\textit{SEDM memory: admission + controller}} \\
No memory & 0.534 & 0.554 & 0\% & 1,476 \\
SEDM, no gate & 0.518 & 0.555 & 100\% & 2,781 \\
SEDM + \cmg{} (val.\ $\lambda{=}0$) & \textbf{0.546} & \textbf{0.564} & 33\% & 1,915 \\
SEDM + \cmg{} ($\lambda{=}0.05$) & 0.537 & 0.557 & 8\% & 1,558 \\
Oracle & 0.596 & -- & -- & -- \\
\bottomrule
\end{tabular}
\caption{\cmg{} inside the full SEDM pipeline on LoCoMo (test conversation 2, 193 questions). \cmg{} rows are means over three seeds. Succ.: SBERT $\ge$ 0.7; F1: token F1; Inj.: injection rate; Tok.: mean agent input tokens per question. The two ``No memory'' rows are independent runs of the same agent.}
\label{tab:locomo}
\end{table}

Always injecting SEDM memory helps little here: with the admission-only memory, success is 0.539 against 0.544 without memory, while agent input tokens rise by 89\%. With the validation-selected $\lambda=0$, \cmg{} injects memory on 65\% of the questions, uses 19\% fewer tokens than always using memory, and reaches 0.556 ($\pm$0.011 over three seeds), above both fixed policies; at $\lambda=0.05$ it reaches 0.568 with 35\% fewer tokens (Table~\ref{tab:locomo}). Token F1 shows the same pattern, and every operating point lies above random gating (Figure~\ref{fig:curve}b,c). Single differences on 193 questions are within the noise of one conversation, but the direction holds for all three seeds, both memories, and both metrics.

The controller-managed memory is the only setting in which always using memory clearly hurts (0.518 vs.\ 0.534). This seems to contradict the gain from scheduling in Table~\ref{tab:efficiency}, but here the memory comes from a single conversation and is frozen, so merged and distilled items become more general but may not fit a new conversation, and the controller cannot correct their weights during testing. This is the case a recipient-side gate is meant for: \cmg{} injects this memory on only 33\% of the questions, saves 31\% of the tokens, and turns the loss of always using it ($-$1.6 points) into a gain ($+$1.2 points). We do not compare the two memories directly, since they come from independent runs.
\vspace{0.6cm}
\subsection{Analysis}
\label{sec:analysis}
\vspace{0.3cm}
\paragraph{Why winner labels fail.} Table~\ref{tab:main} doubles as an ablation of \cmg{}: without outcome supervision (winner head, single-token SFT) or with retrieval similarity alone, the gate collapses to always using memory, and without the single-pass output (CoT gate) it becomes slower and less accurate. Only 22\% of the training tasks are non-ties, and whether a task lands in 01 or 10 is noisy: the winner-label gates reach a validation cross-entropy of 0.689--0.696, which is chance level ($\ln 2 \approx 0.693$). \cmg{} learns from all 2,962 tasks, and its validation loss (1.078--1.080) is below the 1.16 of a constant base-rate predictor.

\paragraph{\cmg{} adapts to memory quality.} In a pilot with LLM-written strategy notes that hurt on average (always use 0.245 vs.\ always skip 0.275; Appendix~\ref{app:pilot}), \cmg{} injected them on only 1\% of the test tasks and matched always skip. With evidence memory, which helps, it injects on 69\%. Across our three main settings, \cmg{} likewise injects the helpful HotpotQA memory most often and the controller-managed LoCoMo memory, which hurts when always used, least often (Figure~\ref{fig:injection} in Appendix~\ref{app:injection}).

\section{Conclusion}
\label{sec:conclusion}

We presented SEDM, a society memory framework that selects what to retain and when to use it. On the sender side, SEDM combines paired A/B admission, utility-aware scheduling, and cross-domain abstraction. Under the original admission configuration, it achieves the highest open-domain F1 among evaluated LoCoMo baselines and improves on G-Memory on FEVER and HotpotQA with fewer prompt tokens. On the recipient side, \cmg{} predicts success with and without memory in one forward pass, using paired outcomes including ties and held-out calibration. It approaches always-use success on HotpotQA with 18\% fewer agent input tokens and achieves higher observed success than both fixed policies in the held-out LoCoMo pipeline, saving 19--31\% relative to always-use. These results support selective use as a complement to selective retention, with an adjustable quality--cost trade-off. Future work will evaluate both as memories evolve in open-ended agent societies \citep{altera2024projectsid}.

\newpage
\section*{Limitations}

\paragraph{Scope.} We evaluate selective memory use on two datasets (HotpotQA and LoCoMo), one agent model (gpt-4o-mini), and one gate backbone. Results may differ for other tasks, agents, and memory types. Our society setting is limited: memories come from other agents' episodes on other tasks or conversations, but we do not yet study large, open-ended societies with many roles.

\paragraph{Admission protocol.} The results in Tables~\ref{tab:locomo-main}--\ref{tab:cross} come from the original SEDM configuration, which tests a candidate on its own source task and may therefore overestimate its utility. We did not rerun them under the stricter held-out configuration; the experiments in Section~\ref{sec:locomo} use the held-out configuration throughout. We ablate admission and scheduling (Table~\ref{tab:efficiency}) but not cross-domain diffusion, so Table~\ref{tab:cross} shows that memory transfers, not how much of the transfer is due to abstraction.

\paragraph{Frozen memory.} In both CMG experiments the memory is frozen during evaluation. We do not study a gate that keeps learning while the society memory keeps changing. The LoCoMo test set is a single conversation, so its confidence intervals are wide.

\paragraph{Memory construction.} In the controlled HotpotQA study, evidence memories use the gold supporting sentences of the memory-construction questions. They model memories written after a past episode was solved and verified, which is an optimistic setting. No test question or its evidence is ever used.

\paragraph{Baselines.} The CoT gate is a zero-shot reproduction with the same backbone, not the original trained checkpoint. Single-pass gate latencies were measured on a shared GPU and vary with load.

\paragraph{Statistics and cost.} The HotpotQA test set has 489 questions, so many differences between gates are within confidence intervals; the comparison with random gating is a point estimate. The token cost in the decision rule is an estimate of the injected memory length; the reported savings are measured API input tokens. Absolute success rates are modest because the open-corpus, seven-step setting is hard; we focus on relative changes.

\section*{Ethical Considerations}
We use public datasets and a commercial API under its terms. Memory gating reduces cost but can also hide useful information from an agent; deployments should monitor the quality loss implied by the chosen $\lambda$.

\bibliography{custom,cmg_extra}

\appendix

\section{SEDM Implementation Details}
\label{app:sedm}

This appendix describes the held-out configuration used in Section~\ref{sec:locomo}; the original configuration is summarized at the end.

\paragraph{Weight initialization and update.}
Every new item starts from the same base weight $w_0$ and is then adjusted by its A/B result:
\begin{equation}
w(m) \leftarrow \mathrm{clip}\big(w_0 + u(m),\, 0,\, w_{\max}\big),
\end{equation}
where the update $u(m)$ depends on the outcome of the test:
\begin{equation}
u(m) =
\begin{cases}
\mathrm{clip}\big(0.5\,\overline{\Delta R},\, -0.3,\, 0.3\big) & \text{if decisive},\\
0.1\,\overline{\Delta R} & \text{otherwise}.
\end{cases}
\end{equation}
A test is decisive when the item is admitted with a positive gain or rejected with a negative gain. A memory that helped therefore starts above the base weight, a memory that hurt starts below it, and a tie keeps the base weight. Because $w_0 > 0$, tied items can still be retrieved and receive further evidence later. The bounded step keeps a single noisy test from deciding the fate of an item. We use $w_0 = 0.1$ and $w_{\max} = 2$.

\paragraph{Controller cycle.}
The controller runs periodically, after a fixed number of processed tasks, and once more at the end of a run. Each cycle has four steps.
(1) \emph{Weight update.} New A/B evidence for an item is applied with the same bounded rule as at initialization, and each update also increases the usage count $n(m)$ of the item.
(2) \emph{Tier management.} Items live in three tiers: short-term, episodic, and long-term. An item moves up one tier when $w(m) > \theta$ and $n(m) \ge n_{\min}$, and reaches the long-term tier when $w(m) > 2\theta$ and $n(m) \ge 2n_{\min}$. It moves down when its weight drops below $\theta/2$, and again below $\theta/4$. We use $\theta = 0.15$ and $n_{\min} = 3$.
(3) \emph{Merge, distillation, and conflict isolation.} Near-duplicate items from different SCECs are merged into one item that keeps their combined support. Patterns that repeat across items are distilled into a new item, which starts from a higher base weight (0.3). A rejected item whose reward change falls below a conflict threshold is isolated: it keeps its record but is never retrieved again.
(4) \emph{Decay.} Weights decay over time,
\begin{equation}
w(m) \leftarrow w(m) \cdot \delta \cdot \tau\big(\Delta t(m)\big),
\end{equation}
where $\delta = 0.98$, $\Delta t(m)$ is the time since the item was last used, and $\tau(\Delta t) = \max\{0.1,\ 1 - \Delta t / 30\,\text{days}\}$. Items whose weight falls below a minimum are removed. Frequently helpful items thus rise to the long-term tier, while unused or harmful items fade out of retrieval.

\paragraph{Admission settings.} In the LoCoMo pipeline we use $e=3$ admission tasks, $k=3$ runs per arm, and $\eta=0$, so that items that do not hurt the admission tasks are also kept.

\paragraph{Original configuration.} The results in Tables~\ref{tab:locomo-main}--\ref{tab:cross} use the original SEDM configuration. There, a candidate is tested on its own source SCEC, admitted if $S(m) \ge \eta$, and initialized with $w_0(m) = \max\{0, S(m)\}$. Retrieval uses $s(q,m) = \mathrm{sim}(q,m) \times w(m)$, and weights are updated as $w_{t+1}(m) = w_t(m) + \alpha\,\bar U_t(m) - \beta\, f_{\mathrm{use},t}(m)$, where $\bar U_t(m)$ is the average realized utility of $m$ since the last update, $f_{\mathrm{use},t}(m)$ its usage frequency, and $\alpha, \beta \ge 0$.

\section{Controlled HotpotQA Study Details}
\label{app:hotpot}

\paragraph{Data and splits.}
We use all 7,405 questions of the HotpotQA distractor development set \citep{yang2018hotpotqa}. Questions are split into five disjoint sets: memory construction (2,962), gate training (2,962), validation (496), calibration (496), and test (489). The test set is used once, for final reporting. A manifest with hashes fixes the split, and every later step checks it.

\paragraph{Memory.}
Memories are built only from the memory-construction split. Each memory records one past episode: the question, its verified answer, and its supporting sentences (\emph{evidence memory}). This gives 2,962 memories with 565 characters on average. A retriever embeds the new question with all-MiniLM-L6-v2 \citep{reimers2019sentencebert,wang2020minilm} and returns the five most similar memories. A check refuses to run if any memory comes from a non-memory split. Memories therefore never contain a training, validation, calibration, or test question; whether they help depends on whether they share entities or facts with the new question.

\paragraph{Agent and execution protocol.}
The agent is gpt-4o-mini-2024-07-18 with temperature 0, at most 256 output tokens per step and at most seven steps. It follows ReAct \citep{yao2023react} with three actions: \texttt{Search[entity]}, \texttt{Lookup[keyword]}, and \texttt{Finish[answer]}. Tools run over a frozen corpus of 66,581 paragraphs: all paragraphs of the development set, merged by title and identical for every task and both actions. \texttt{Search} returns the best lexical match by title and text. Under \USE{}, the five memories are appended to the prompt under ``Retrieved memories''; this is the only difference between the two runs. A task succeeds if the agent calls \texttt{Finish} within the step limit and its answer has token F1 $\ge 0.5$ with the reference (HotpotQA normalization). We stop generation at ``\texttt{\textbackslash nObservation}'' and execute only the first action of each turn, so the agent cannot invent its own tool results (Appendix~\ref{app:protocol}). Runs are cached by a hash of the task, action, memories, agent configuration, and corpus.

\paragraph{Gate training.}
All learned gates use Qwen3-4B with LoRA (rank 16, $\alpha=32$, on query, key, value, and output projections), learning rate $10^{-4}$, effective batch size 16, three epochs, bf16, and gradient checkpointing on one NVIDIA A40. We keep the epoch with the lowest validation loss. \cmg{} is trained with three seeds (42, 43, 44); other gates use seed 42.

\paragraph{Baselines.}
All gates are scored on the same paired test executions.
\textbf{Always skip} and \textbf{always use} are the two ends of the decision space.
\textbf{Similarity threshold} uses memory if the highest retrieval similarity exceeds $\tau \in \{0.2,\dots,0.6\}$, chosen on validation.
\textbf{Random gating} uses memory with probability $u$; its expected success at injection rate $u$ is $(1-u)\,S_{\SKIP} + u\,S_{\USE}$.
\textbf{Winner head} is the same model with a two-way softmax head trained with cross-entropy on non-tie tasks only.
\textbf{Single-token SFT} reads the language-model logits of the next tokens ``skip'' and ``use'' and is trained with the same winner labels; it removes CoT but keeps the old supervision.
\textbf{CoT gate} is a generative gate that writes a chain of thought before deciding, run zero-shot with the same backbone (Qwen3-4B, thinking mode off, greedy decoding, at most 384 new tokens). It must end with ``Decision: USE'' or ``Decision: SKIP''. All 489 test outputs could be parsed.
\textbf{Oracle} picks a successful action whenever one exists.

\paragraph{Metrics.}
We report success rate with 95\% bootstrap confidence intervals (2,000 resamples over questions; \citealp{efron1994bootstrap}), injection rate (share of tasks with \USE{}), mean agent input tokens per task, and gate latency (median per decision, batch size 1, after one warm-up, including tokenization). Output tokens per task barely change across policies (142--149) and are omitted. For calibration we report NLL, Brier score, and ECE with ten equal-width bins.

\paragraph{Calibration.}
All fitted temperatures are above one (1.13--1.49), so the raw heads are overconfident (Table~\ref{tab:calib}). Temperature scaling lowers the ECE of the \USE{} head from 0.060 to 0.044 and slightly improves NLL and Brier score for both heads. With one seed, calibrated and raw decisions at $\lambda=0$ score almost the same. Across seeds, however, calibration matters: the raw gate ranges from 0.305 to 0.339, while the calibrated gate stays between 0.323 and 0.329. Because the decision rule compares a probability difference with a cost, well-scaled probabilities make the same $\lambda$ mean the same thing across training runs.

\begin{table}[h]
\centering
\small
\setlength{\tabcolsep}{4pt}
\begin{tabular}{@{}lccc@{}}
\toprule
Head & NLL & Brier & ECE \\
\midrule
$\pskip$ & 0.543 $\to$ 0.538 & 0.177 $\to$ 0.177 & 0.051 $\to$ 0.049 \\
$\puse$ & 0.584 $\to$ 0.579 & 0.198 $\to$ 0.196 & 0.060 $\to$ 0.044 \\
\bottomrule
\end{tabular}

\vspace{4pt}
\resizebox{\columnwidth}{!}{%
\begin{tabular}{@{}lccc|c@{}}
\toprule
Seed & 42 & 43 & 44 & Mean $\pm$ sd \\
\midrule
$T_{\SKIP}$ / $T_{\USE}$ & 1.16 / 1.13 & 1.28 / 1.42 & 1.49 / 1.22 & -- \\
Raw, $\lambda=0$ & 0.327 & 0.305 & 0.339 & 0.324 $\pm$ 0.017 \\
Calibrated, $\lambda=0$ & 0.329 & 0.323 & 0.325 & 0.326 $\pm$ 0.003 \\
Val.-selected $\lambda$ & 0.329 & 0.301 & 0.325 & 0.318 $\pm$ 0.015 \\
\bottomrule
\end{tabular}}
\caption{Top: test calibration of \cmg{} before $\to$ after temperature scaling (seed 42). Bottom: fitted temperatures and test success rate across seeds. For seed 43, validation selected $\lambda=0.05$.}
\label{tab:calib}
\end{table}

\section{LoCoMo Pipeline Details}
\label{app:locomo}

\paragraph{Splits.} LoCoMo contains ten conversations. We split by conversation so that no conversation is shared between roles: conversation 0 produces candidate memories, conversation 1 is used for admission, conversations 3--7 provide paired training data for \cmg{} (1,089 questions), conversation 8 is used for validation, conversation 9 for calibration, and conversation 2 for testing (193 questions).

\paragraph{Pipeline.} Agents answer with the SEDM pipeline (AutoGen with G-Memory retrieval). Admission uses the held-out configuration (\S\ref{sec:admission}) with $e=3$, $k=3$, $\eta=0$, $\lambda_L=\lambda_T=0$, and both filters. The controller-managed memory additionally updates weights, moves memories between tiers, merges near-duplicates, distills patterns, and decays unused items. Each memory is frozen before gate training, and \cmg{} is trained separately for each memory.

\paragraph{Execution and metrics.} Every question is executed twice through the full pipeline, with the retrieved SEDM memories forced off and on. The agent is gpt-4o-mini. Following SEDM, a question counts as solved if the SBERT similarity between the answer and the reference is at least 0.7; we also report token F1.

\section{Injection Rate across Memories}
\label{app:injection}

Figure~\ref{fig:injection} shows how often \cmg{} injects memory at each price $\lambda$ in the three settings of Sections~\ref{sec:cmg-exp} and~\ref{sec:locomo}.

\begin{figure}[h]
\centering
\begin{tikzpicture}
\begin{axis}[
  width=\columnwidth, height=0.62\columnwidth,
  xlabel={Price $\lambda$}, ylabel={Injection rate (\%)},
  symbolic x coords={0,0.05,0.1,0.2,0.5,1}, xtick=data,
  ymin=0, ymax=85,
  tick label style={font=\scriptsize}, label style={font=\small},
  legend style={font=\scriptsize, at={(0.98,0.97)}, anchor=north east, draw=none, fill=white, fill opacity=0.85, text opacity=1},
  grid=major, grid style={gray!20},
]
\addplot[blue!70!black, thick, mark=*, mark size=1.5pt] coordinates {(0,76.0) (0.05,47.0) (0.1,27.0) (0.2,12.3) (0.5,1.7) (1,0)};
\addlegendentry{HotpotQA (memory: $+$8.4 pts)}
\addplot[orange!80!black, thick, mark=square*, mark size=1.5pt] coordinates {(0,65.3) (0.05,30.0) (0.1,3.0) (0.2,0) (0.5,0) (1,0)};
\addlegendentry{LoCoMo, adm.\ only ($-$0.5 pts)}
\addplot[red!70!black, thick, mark=triangle*, mark size=2pt] coordinates {(0,33.3) (0.05,7.7) (0.1,0) (0.2,0) (0.5,0) (1,0)};
\addlegendentry{LoCoMo, adm.\ + ctrl.\ ($-$1.6 pts)}
\end{axis}
\end{tikzpicture}
\caption{Injection rate of \cmg{} as the price $\lambda$ grows (test splits, mean of three seeds). The legend gives the effect of always using memory on the test split. The more a memory helps, the more often \cmg{} injects it at every price.}
\label{fig:injection}
\end{figure}

\section{Execution Protocol Details}
\label{app:protocol}

\paragraph{Agent prompt.} The system prompt asks the agent to answer with \texttt{Finish[short answer]}, to take exactly one action per turn, never to write an observation itself, and it states that retrieved memories are fallible reference material. \texttt{Search} returns the first paragraph of the best-matching document; \texttt{Lookup} returns the next paragraph of the current document that contains the keyword.

\paragraph{Invented observations.} Without a stop sequence, the agent sometimes writes a tool call, then an invented ``Observation'', and then a final answer in the same turn. Earlier versions of our pipeline executed such answers, so tool results were partly imagined. We stop at ``\texttt{\textbackslash nObservation}'' and keep only the first action. Over all runs, text after the first action was discarded 331 times on the training split, 63 on validation, 53 on calibration, and 58 on test.

\paragraph{Gate input.} The gate reads a JSON object with the question and the list of memory texts, preceded by ``Predict task success without and with the retrieved memories.'' and followed by ``Decision (skip or use):''. Answers, task identifiers, scores, and traces are excluded.

\paragraph{Cost.} Collecting all paired runs used about 24.7M input and 1.3M output tokens of gpt-4o-mini (about US\$4.5).

\section{Paired Label Statistics}
\label{app:labels}

Table~\ref{tab:labels} lists the paired outcomes of the controlled HotpotQA study for each split. Ties (00 and 11) make up most tasks in every split.

\begin{table}[!ht]
\centering
\small
\resizebox{\columnwidth}{!}{%
\begin{tabular}{@{}lrrrrrr@{}}
\toprule
Split & $n$ & $S_{\SKIP}$ & $S_{\USE}$ & 00 & 01 & 10 / 11 \\
\midrule
Train & 2,962 & 0.249 & 0.307 & 1,814 & 409 & 239 / 500 \\
Validation & 496 & 0.244 & 0.288 & 312 & 63 & 41 / 80 \\
Calibration & 496 & 0.254 & 0.310 & 300 & 70 & 42 / 84 \\
Test & 489 & 0.249 & 0.333 & 294 & 73 & 32 / 90 \\
\bottomrule
\end{tabular}}
\caption{Paired outcomes per split. 01: only \USE{} succeeds; 10: only \SKIP{} succeeds.}
\label{tab:labels}
\end{table}

\section{Full $\lambda$ Curves for All Seeds}
\label{app:seeds}

Table~\ref{tab:seeds} reports the test success and injection rate of calibrated \cmg{} in the controlled HotpotQA study for every price $\lambda$ and all three seeds.

\begin{table}[!h]
\centering
\small
\setlength{\tabcolsep}{4pt}
\begin{tabular}{@{}lccc@{}}
\toprule
$\lambda$ & Seed 42 & Seed 43 & Seed 44 \\
\midrule
0 & 0.329 / 69\% & 0.323 / 82\% & 0.325 / 77\% \\
0.05 & 0.315 / 40\% & 0.301 / 45\% & 0.317 / 56\% \\
0.1 & 0.288 / 24\% & 0.282 / 19\% & 0.305 / 38\% \\
0.2 & 0.282 / 13\% & 0.270 / 5\% & 0.290 / 19\% \\
0.5 & 0.262 / 3\% & 0.249 / 0\% & 0.260 / 2\% \\
1 & 0.249 / 0\% & 0.249 / 0\% & 0.249 / 0\% \\
\bottomrule
\end{tabular}
\caption{Test success / injection rate of calibrated \cmg{} for each $\lambda$ and seed. The shape of the curve is stable; the injection rate at a given $\lambda$ varies across seeds, so gates are best compared at matched injection rates.}
\label{tab:seeds}
\end{table}

\section{Pilot Run with Insight Memory}
\label{app:pilot}

The pilot used a random 2,000-question subset split into memory construction (400), training (1,000), validation (200), calibration (200), and test (200). Each memory was a strategy note of at most 60 words, written by gpt-4o-mini after it answered a memory-construction question without memory. All other settings follow Section~\ref{sec:exp}.

\begin{table}[h]
\centering
\small
\begin{tabular}{@{}lccc@{}}
\toprule
Gate & Success & Injection & Latency (s) \\
\midrule
Always skip & 0.275 & 0\% & -- \\
Always use & 0.245 & 100\% & -- \\
Winner head & 0.270 & -- & -- \\
Single-token SFT & 0.270 & -- & -- \\
CoT gate & 0.270 & 5\% & 2.70 \\
\cmg{}, $\lambda=0$ & 0.275 & 1\% & 0.06 \\
\bottomrule
\end{tabular}
\caption{Pilot results (200 test questions, seed 42). \cmg{} seeds 42/43/44: 0.275 / 0.275 / 0.255.}
\end{table}

\section{Reproducibility}
Code for data preparation, paired collection, gate training, calibration, and evaluation will be released. Every reported number is generated by one evaluation script from the saved predictions.

\end{document}